\documentclass[11pt,a4paper]{article}
\usepackage[]{acl}
\usepackage{times}
\usepackage{latexsym}
\usepackage{multirow}

\usepackage{times}
\usepackage{latexsym}
\usepackage{cancel}
\usepackage{graphicx}
\usepackage{amsmath}
\usepackage{latexsym}
\usepackage{amssymb}
\usepackage[normalem]{ulem}
\usepackage{amsthm}
\usepackage{color}
\usepackage{xspace}
\usepackage{subfigure}
\usepackage{booktabs}

\usepackage{array, booktabs, makecell}

\usepackage{booktabs}
\usepackage{array,tabularx}
\usepackage[ruled,vlined]{algorithm2e}
\usepackage{algpseudocode}
\usepackage{amsmath}
\usepackage{amsfonts}
\usepackage{latexsym}
\usepackage{xspace}
\usepackage{booktabs}
\usepackage{multirow}

\usepackage{microtype}
\usepackage[ruled,vlined]{algorithm2e}
\usepackage{algpseudocode}
\usepackage{microtype}

\usepackage{xcolor}	
\usepackage{colortbl}
\newcommand{\eat}[1]{}
\newcommand{\red}[1]{\textcolor{red}{#1}}

\title{Learning to Repair: Repairing model output errors after deployment using a dynamic memory of feedback}

\author{Niket Tandon\thanks{\hspace{0.5em}Equal Contribution}\hspace{0.5em}, Aman Madaan~\footnotemark[1]\hspace{0.5em}$^\dagger$,  Peter Clark,  Yiming Yang$^\dagger$,\\ 
  Allen Institute for Artificial Intelligence, Seattle, WA, USA \\ 
  $^\dagger$ Language Technologies Institute, Carnegie Mellon University, Pittsburgh, PA, USA \\
  \texttt{\{nikett,peterc\}@allenai.org} \\ \texttt{\{amadaan,yiming\}@cs.cmu.edu} }

\date{}
\usepackage{algorithm2e}
\usepackage{pgfplots}
\pgfplotsset{compat=1.17}  

\definecolor{Red}{rgb}{1,0,0}
\definecolor{Green}{rgb}{0.4,1,0.2}
\definecolor{Blue}{rgb}{0,0,1}
\definecolor{Red}{rgb}{0.9,0,0}
\definecolor{Orange}{rgb}{1,0.5,0}
\definecolor{cosmiclatte}{rgb}{1.0, 0.97, 0.91}
\definecolor{yellow}{rgb}{0.65,0.6,0}
\definecolor{cadmiumgreen}{rgb}{0.2, 0.7, 0.24}
\definecolor{cornellred}{rgb}{0.7, 0.11, 0.11}

\newcommand{\V}[1]{\mathbf{#1}}

\newcommand{\green}[1]{\textcolor{cadmiumgreen}{#1}}

\newcommand{\secref}[1]{\S\ref{#1}}

\newcommand{\proscript}{\textsc{proscript}\xspace}
\newcommand{\proscriptgen}{\textsc{proscript}$_{gen}$\xspace}
\newcommand{\ie}{i.e.,\xspace}
\newcommand{\eg}{e.g.,\xspace}
\newcommand{\iset}{\textit{interaction-reuse set}}
\newcommand{\isetMother}{\iset-genesis\xspace}

\newcommand{\ncr}[1]{#1}

\newcommand{\X}[1]{\mathbf{\mathrm{S}}}
\newcommand{\Z}[1]{\mathbf{\mathrm{C^-}}}
\newcommand{\VV}[1]{\mathbf{\mathrm{C^+}}}
\newcommand{\W}[1]{\mathbf{\mathrm{M^-}}}
\newcommand{\U}[1]{\mathbf{\mathrm{S^-}}}
\newcommand{\Y}[1]{\mathbf{\mathrm{M^+}}}
\newcommand{\LL}[1]{\mathbf{\mathrm{H^-}}}
\newcommand{\M}[1]{\mathbf{\mathrm{H^+}}}

\newcommand{\nfb}{\textsc{no-fb}}

\newcommand{\iscript}{\textsc{interscript}\xspace} 
\newcommand{\oursnom}{\textsc{FBNet}\xspace} 
\newcommand{\ours}{$\oursnom$\xspace}
\newcommand{\oursnomo}{\textsc{FBNet}$_o$\xspace} 
\newcommand{\oursoraclefb}{$\oursnomo$\xspace}
\newcommand{\badgm}{$y_{bad}$\xspace}

\newcommand{\fbm}{$\V{fb}$\xspace}

\newcommand{\dotf}{\textsc{dot}\xspace}

\newcommand{\ques}{\V{x}}
\newcommand{\ans}{\V{y}}
\newcommand{\basemodel}{\V{B}}

\newcommand{\ret}{\Omega}
\newcommand{\memory}{\mathcal{M}}
\newcommand{\errorstate}{\V{e}}
\newcommand{\fb}{\V{fb}}
\newcommand{\corrector}{\V{G}}
\newcommand{\edit}{\V{y}^e}

\newcommand{\quesm}{$\ques$\xspace}
\newcommand{\ansm}{$\ans$\xspace}
\newcommand{\basemodelm}{$\basemodel$\xspace}
\newcommand{\retm}{$\ret$\xspace}
\newcommand{\memorym}{$\memory$\xspace}
\newcommand{\errorstatem}{$\errorstate$\xspace}
\newcommand{\correctorm}{$\corrector$\xspace}

\definecolor{bluetype}{HTML}{15616D}
\definecolor{bluearg}{HTML}{CA2E55}
\definecolor{orangeloc}{HTML}{190E4F}

\newcommand{\editm}{$\edit$\xspace}
\newcommand{\editgoldm}{$\V{y}^{e^*}$\xspace}
\newcommand{\editgeneratedm}{$\V{y}^{\hat{e}}$\xspace}
\newcommand{\edittype}[1]{\textcolor{bluetype} {#1}}
\newcommand{\editarg}[1]{\textcolor{bluearg}{#1}}
\newcommand{\editloc}[1]{\textcolor{orangeloc}{#1}}

\newcommand{\tfxxl}{\textsc{t5-xxl}\xspace}

\newcommand{\exact}{\texttt{EM}\xspace}
\newcommand{\bleu}{\texttt{BLEU}\xspace}
\newcommand{\rouge}{\texttt{ROUGE}\xspace}

\def\@withdot.{\ifmmode\!\string/\!
               \else\kern-1.8pt\string/\kern-1.8pt\fi.}

\newcommand{\squishlist}{
  \begin{list}{$\bullet$}
    { \setlength{\itemsep}{0pt}      \setlength{\parsep}{3pt}
      \setlength{\topsep}{3pt}       \setlength{\partopsep}{0pt}
      \setlength{\leftmargin}{1.5em} \setlength{\labelwidth}{1em}
      \setlength{\labelsep}{0.5em} } }
\newcommand{\reallysquishlist}{
  \begin{list}{$\bullet$}
    { \setlength{\itemsep}{0pt}    \setlength{\parsep}{0pt}
      \setlength{\topsep}{0pt}     \setlength{\partopsep}{0pt}
      \setlength{\leftmargin}{0.2em} \setlength{\labelwidth}{0.2em}
      \setlength{\labelsep}{0.2em} } }

 \newcommand{\squishend}{
     \end{list} 
 }

\begin{document}
\maketitle

\begin{abstract}

Large language models (LMs), while powerful, are not immune to mistakes, but can be difficult to retrain. Our goal is for an LM to continue to improve after deployment, without retraining, using feedback from the user. Our approach pairs an LM with (i) a growing memory of cases where the user identified an output error and provided general feedback on how to correct it (ii) a {\it corrector model}, trained to translate this general feedback into specific edits to repair the model output.
Given a new, unseen input, our model can then use feedback from similar, past cases to repair output errors that may occur. We instantiate our approach using an existing, fixed model for {\it script generation}, that takes a goal (e.g., ``bake a cake'') and generates a partially ordered sequence of actions to achieve that goal, sometimes containing errors. Our memory-enhanced system, \ours, learns to apply user feedback to repair such errors (up to 30 points improvement), while making a start at avoiding similar past mistakes on new, unseen examples  (up to 7 points improvement in a controlled setting). This is a first step towards strengthening deployed models, potentially broadening their utility.\footnote{Our code and data is available at \url{https://github.com/allenai/interscript}
}
\end{abstract} 


\section{Introduction}

\begin{figure}[!t]
\centering
     {\includegraphics[width=1.05\columnwidth]{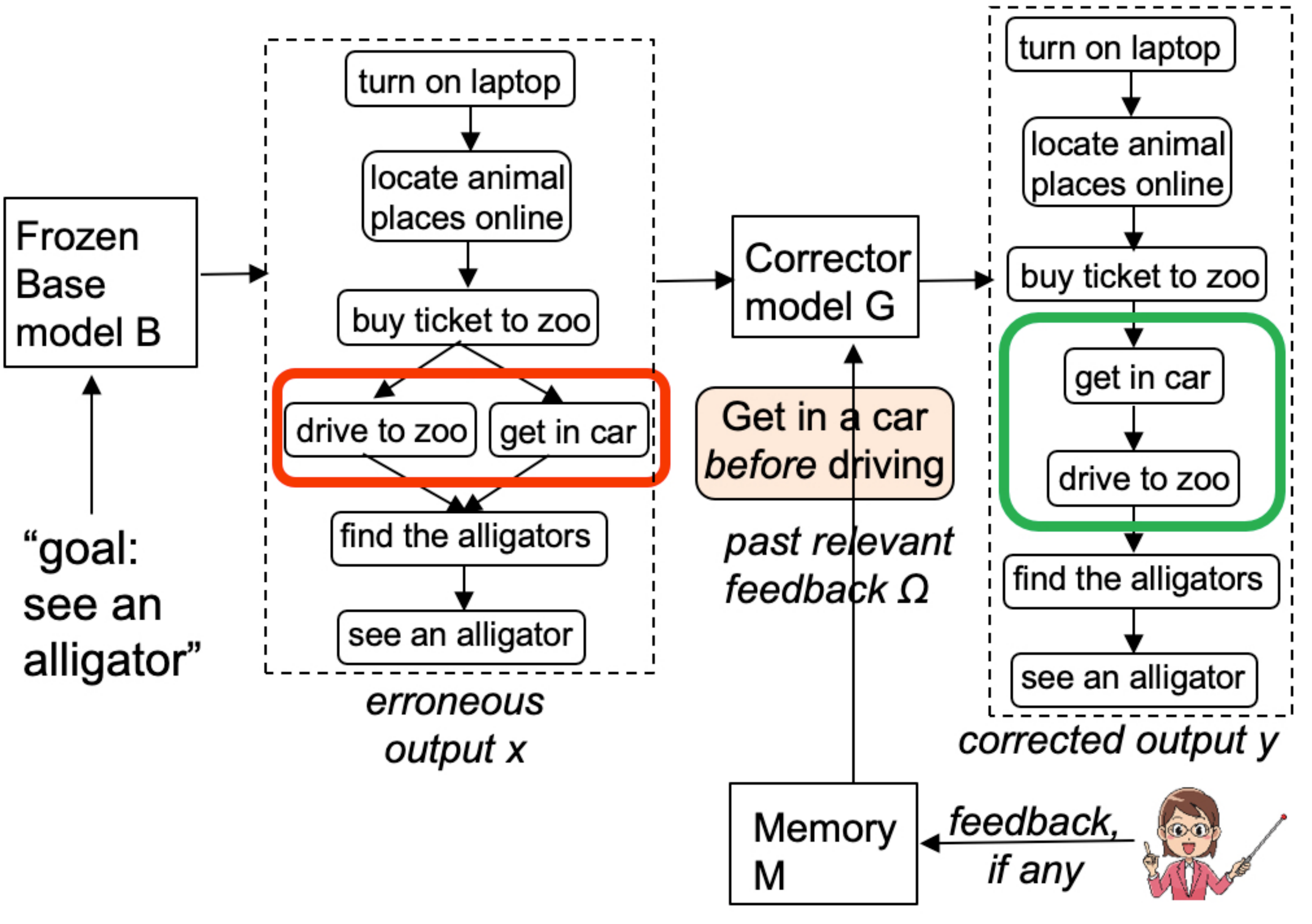}}
   \caption{
     Given a frozen model \basemodelm, we train a {\it corrector model} \correctorm to
     apply feedback from a user about errors made by the original model. In the example,
     \basemodelm has generated a script with an error in, stating that ``driving'' and
     ``getting in a car'' can occur in any order (red box). The user provides general feedback
     (``Get in a car before driving''), and \correctorm operationalizes this to
     generate a corrected graph (by predicting and applying a graph edit operation) in which ``get in car'' happens first (green box). The feedback is stored in a memory \memorym so it can also be retrieved to repair similar, future errors. }
\label{fig:intro-example}
\vspace{-3mm}
\end{figure}

Language models (LMs) have achieved remarkable success on many tasks
\cite{Wang2018GLUE, talmor-etal-2019-commonsenseqa}, but they are still prone to mistakes \citep{BenderKoller2020FormAndMeaning}. Correcting mistakes by retraining is not always easy due to the cost and/or unpredictability of how additional training data will change the model. Instead, our goal is to allow users to correct such errors directly through interaction,  without retraining -- by giving corrective feedback on the model's output.
Our approach is to maintain a growing, dynamic memory of such feedback,
and use a trained {\it corrector model} to apply such feedback to
repair the model output. By doing so, the system can also potentially fix
output errors for new unseen inputs using feedback from similar, past cases.
\ncr{The ability to leverage a fixed trained model without re-training could save costs and have a positive environmental impact.} 

We consider the class of problems where the model's output is {\it repairable},
namely a structured output that is (typically) nearly correct,
and fixable through a small number of edit operations. \ncr{Our system is general and admits a general graph based input, so in principle it applies to a large number of tasks. In this paper, we apply our approach to the task of {\it script generation} that provides a natural setting for users to critique, and has applications in smart assistants \cite{learning-decompose-tasks-2021}. We use an existing, fixed model: proScript \cite{proscript} 
that satisfies the constraint of the model's output to be repairable.} proScript takes as input a goal to achieve (expressed in natural language), and outputs a
partially ordered sequence of steps - a {\it script} - required to achieve
that goal. Our interest here is not in proScript itself, but in what to do
when proScript's output contains an error.

This instantiation of our approach is illustrated in Figure~\ref{fig:intro-example}. Here, proScript has generated a script \quesm to achieve the goal ``see an alligator'', but the script contains an error: it states that the steps of ``driving to the zoo''
and ``get in car'' can be applied in any order. To repair this, the user provides
the general feedback ``Get in a car before driving''.
The corrector model \correctorm then takes that feedback and the
erroneous script, translates it into appropriate edit operations on the script,
and applies those edits to generate a corrected script (\ansm in Figure~\ref{fig:intro-example}). The feedback is stored in memory \memorym so it can also be retrieved in the future. Our system, \ours, comprises the corrector module \correctorm, the memory \memorym, and searching and writing operations. 
To train our system, we collect examples of bad outputs, general feedback, and
specific edits that the feedback should translate to (Section~\ref{dataset-collection}). 
This allows \correctorm to learn how to translate general feedback into specific edits to apply. Pairing \correctorm with the memory \memorym allows \ours to repair new, unseen scripts containing similar errors to the one the user corrected.

Our approach loosely follows some early AI systems that maintained a memory of the output
problems and how to fix them~\citep{Sussman1973ACM, Hammond1986CHEFAM, Riesbeck1981FailureDrivenRF},
but here, we use neural methods and interact with a user to provide corrective feedback.
It also builds on the idea of allowing users to specify edits in
natural language, e.g., NLEdit \cite{elgohary2021nl}, except we
use {\it general} user feedback (then translated to example-specific edits by \correctorm)
and add a memory so that feedback can also be automatically reused.

We evaluate \ours along two dimensions:
(a) How well does \ours interpret NL feedback?
(b) How well can \ours learn from prior mistakes?
We find that (a) it uses NL feedback effectively to repair script errors,
with +30\% (absolute) improvement over a baseline that does not use feedback,
and that (b) it makes a start at avoiding past mistakes (+7\% (absolute) improvement in a controlled setting).
Although these results are only for a single deployment of our general approach,
they suggest that memory-based architectures can help deployed models
continue to improve with time, without retraining, potentially broadening
their utility.

\section{Related work}

There have been numerous approaches to using user feedback to improve model performance, including: \\
(1) {\bf Providing additional training examples:} \citet{Dasgupta2019MachineTeaching} show how a user can correct bad model behavior by carefully selecting new training examples for the system to learn from, a style of interactive active learning \cite{Settles2012ActiveL}. \\
(2) {\bf Marking/scoring the system's answer(s):} In SHRDLURN, the user provides feedback
  by identifying which of the system's alternative interpretations of a user command is correct \cite{percy-2016-learning-language-games-interaction-shouldrn}. \\
(3) {\bf Providing hints:} \cite{Mehta2019interactionRobotUsingAdvice} show how a system can learns to understand regional (e.g., ``top left") and
  directional (e.g., ``move down") hints from the user for a (simulated) robot. \\
(4) {\bf Provide additional information:} In TeachYourAI \cite{Talmor2020TeachingPM}, given a wrong answer to a question, users can
   enter NL facts and rules to use as context when reasking the question, to (ideally) produce the correct answer. \\
(5) {\bf Correcting bad answers:} In the semantic parsing task of NL-to-SQL, NLEdit learns to interpret and apply
   syntactic edit operations from the user expressed in NL, e.g., ``replace course id with program id." \cite{elgohary2021nl}. \\

These methods all augment/replace the standard use of automated answer feedback (if available), e.g., testing whether
a semantic parse correctly executes to the correct answer, e.g., \cite{Zettlemoyer2005LearningTM}, sometimes using
unsupervised techniques to generate additional training data, e.g., BIFI \cite{yasunaga2021break}.

Our work expands on the above approaches in two important ways.
First, users provide {\it general} feedback in NL, that can potentially be applied to {\it multiple} cases (rather than just
correcting a specific instance). The corrector model \correctorm is trained to operationalize that
advice in different ways for different examples appropriately, in contrast to (say) NLEdit where the user-provided
specific corrective edits for a single example only.

Second, we use a feedback memory, allowing feedback to be reused. While adding external memory to neural systems
is not new, e.g., RAG \cite{Lewis2020RetrievalAugmentedGF}, REALM \cite{Guu2020REALMRL}, ours is the first to utilize a memory of prior user feedback
to improve future neural model performance. This can be viewed as a modern approach to failure-driven reminding,
an essential theme in earlier AI and Cognitive Science research \cite{Riesbeck1981FailureDrivenRF,Schank1989CreativityAL,Ross1984RemindingsAT}.

\section{\ours}
\label{sec:method}

\subsection{Overview of the Architecture}

\begin{figure}[t]
\centering
    \includegraphics[width=\columnwidth]{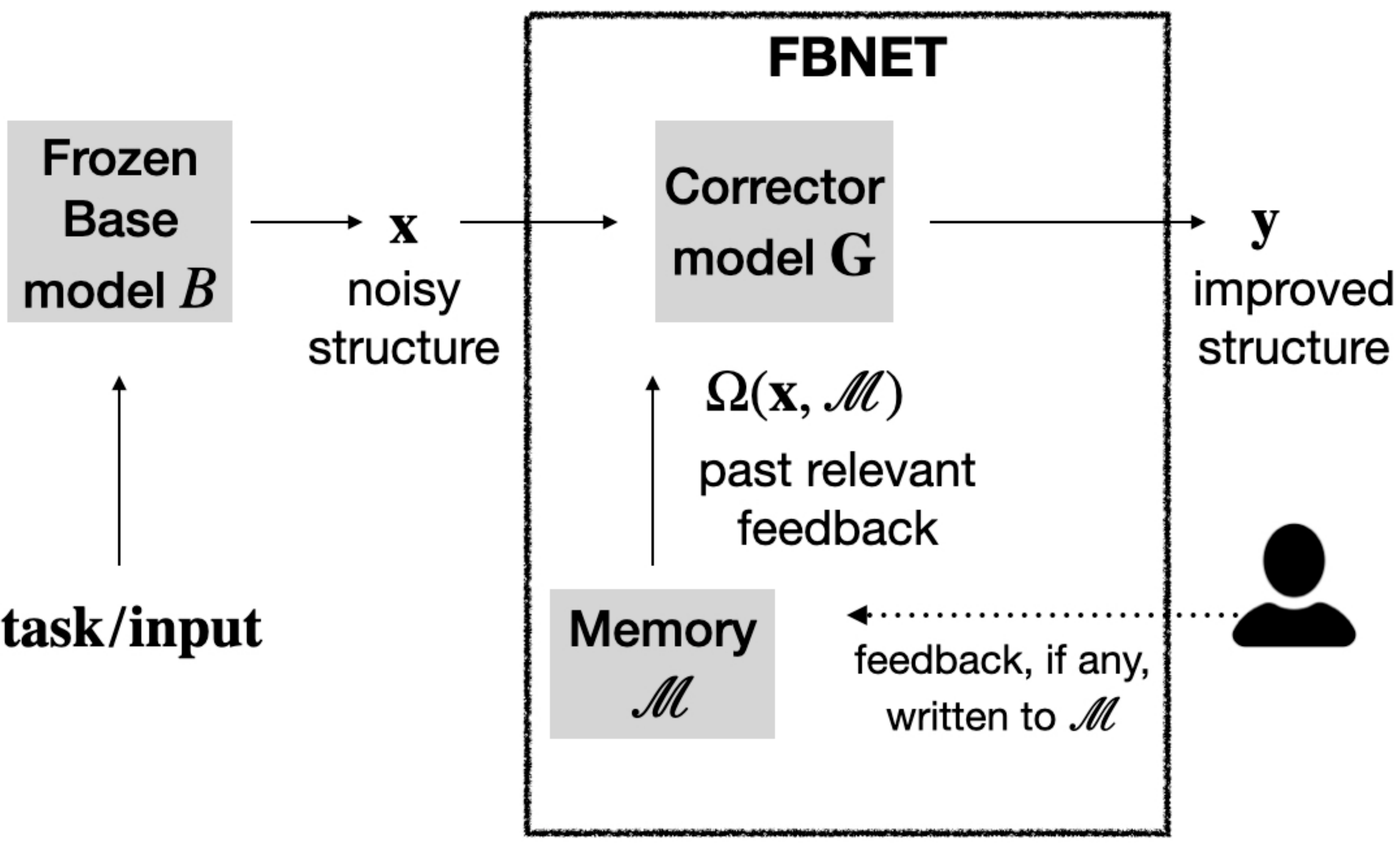}
    \caption{Proposed architecture: (left) \basemodelm does not account for user feedback. (right) \ours maintains a memory \memorym of corrective feedback, and searches for feedback from prior queries with similar error intent as $\ques$ using a retrieval function \retm. $\ques$ is then concatenated to the retrieved feedback to form the input to the corrector model \correctorm. Users can also give new feedback which is added to $\memory$. \ncr{In this work, we focus on script generation models that might generate an erroneous script which are correctable using an edit (feedback).}}
    \label{fig:method}
\end{figure}

Fig. \ref{fig:method} gives an overview of \ours. 
The input is a potentially noisy graph \quesm generated by a base model \basemodelm and the output \ansm is a corrected graph. At inference time, \ie after deployment, a user can critique \ansm by providing natural language feedback \fbm on an error \errorstatem. As output, the model generates the corrected graph \ansm that accounts for \fbm.

The corrector model \correctorm is responsible for improving the potentially noisy output from \basemodelm. \correctorm achieves it using user feedback stored in a continuously updated memory \memorym.

The Memory \memorym is a growing lookup table of key-value pairs: key~($\ques_i$) - value~($\V{fb}_i$), where $\ques_i$ is a particular incorrect graph, and $\V{fb}_i$ is the corresponding feedback. This memory supports lookup (read) and write operations.
Given a new query \quesm, \ours uses feedback \fbm from similar, prior queries in the memory to enrich \quesm.
This feedback \fbm is retrieved using the lookup function $\Omega(\ques, \mathcal{M})$. 
The corrector then combines \fbm with \quesm, and generates \ansm.
The write operation is used whenever a user gives new feedback.

\subsection{Assumptions}
\label{sec:assumptions}
We make two assumptions on the characteristics of the feedback and the input.
\squishlist
\item[A1.] Base model \basemodelm's output is \textbf{repairable}: \basemodelm typically produces syntactically correct output graph but can have semantic errors that the user can recognize and describe using natural feedback. \ncr{For example, the script in Figure \ref{fig:intro-example} is repairable.}

\item[A2.] Feedback is \textbf{reusable}: If two examples $i, j$ have similar errors $e_i$ and $e_j$  then the feedback $fb$ for one should apply to the other, i.e., $(e_i \sim e_j \Leftrightarrow fb_i \sim fb_j)$ 
\squishend

\subsection{Memory \memorym and \retm}
As mentioned, the feedback is stored in a memory of key~($\ques$), value~($\V{fb}$) pairs. 
\retm is a retrieval function that matches a query key~($\ques_j$) to a similar $\ques_i$ in memory implicitly on the similarity of the errors $e_i$ and $e_j$.

\subsection{Corrector model \correctorm}

The graph corrector model \correctorm generates an improved output \ansm given a noisy graph \quesm and \fbm. 
This is done in a two-step process, (i) learning to predict a graph edit operation \editm given  \quesm and \fbm (ii) using simple graph operations to apply \editm to \quesm to produce \ansm.
Our approach of generating an edit instead of directly generating the corrected graph is beneficial for two reasons. First, generating edits is simpler for the model than generating entire graphs.
Second, it simplifies evaluation metrics as it is much simpler to compare two smaller generated edits.
Note that we can deterministically fix a script given an edit. Thus, the two-step process helps us achieve the same end goal (corrected scripts from noisy scripts and feedback).

\subsection{Training and Inference}
\label{sec:training-implementation}
As mentioned, the graph corrector $\V{G}$ first generates an edit \editm, which is applied to the incorrect graph $\ques$ to generate the correct graph $\ans$.
We need a corpus of $(\ques,\fb,\ans)$ to train this system. 
Specifically, we extract an edit from each such tuple, where edit $y^e$ is the difference between the output $\ans$ and the input $\ques$. \quesm and \ansm can be expressed in a string representation using a graph description language such as \dotf. 
We then train a language model to estimate $P_{\theta}(y^e \mid \ques, \fb)$, which allows us to generate an edit for a given $(\ques, \fb)$ using greedy sampling, where $\theta$ denotes the parameters of the language model.

\section{Application: Script Generation}

\begin{table*}[!h]
    \centering
    \small
    \begin{tabular}{|p{0.10\textwidth}|p{0.25\textwidth}|p{0.12\textwidth}|p{0.12\textwidth}||p{0.12\textwidth}|p{0.03\textwidth}|p{0.05\textwidth}|p{0.05\textwidth}|}
    \hline 
    \textbf{Error type} &\textbf{Input script \quesm} &  \textbf{Feedback \fbm} & \textbf{Expected edit \editgoldm}& \textbf{Generated edit \editgeneratedm}& & \textbf{score} & \\ 
    &&&&&$EM$ & $EM_{type}$& $EM_{loc}$\\ \hline
    
    missing step
    & 
    \reallysquishlist 
        \item[] 1. get out of car 
        \item[] 2. stop in front of car
        \item[] 3. turn body toward back of car
        \item[] 4. \red{walk to back of car}
        \item[] 5. \red{take blanket out of car}
        \item[] 6. walk to desired location 
        \item[] 7. throw blanket down 
    \squishend
    & 
    a person needs to open the door before they take an object out	
    &
    \edittype{insert node} \editarg{`open the back door of the car'} \editloc{before `take blanket out of car'}	
    &
    \edittype{insert node} \editarg{`open car door'} \editloc{before `take blanket out of car'}
    &
    0 & 1 & 1
    \\ \hline

    missing step
    & 
    \reallysquishlist 
        \item[] 1. buy a video game 
        \item[] 2. talk to the cashier
        \item[] 3. make the transaction
        \item[] 4. \red{get the receipt}
        \item[] 5. \red{load video game into the car }
        \item[] 6. get into the car 
        \item[] 7. take xbox home 
    \squishend
    &
    after a person makes a transaction, they then head to their car	
    &
    \edittype{insert node} \editarg{`walk to the car'} \editloc{after `get the receipt'}	
    &
    \edittype{insert node} \editarg{`get into the car'} \editloc{after `make the transaction'}
    &
    0 & 1 & 0
    \\ \hline

    wrong step
    & 
    \reallysquishlist 
        \item[] 1. make a bunch of cards
        \item[] 2. grab a pen
        \item[] 3. grab some paper
        \item[] 4. \red{pick up a pen}
        \item[] 5. place the paper on the table
        \item[] 6. \red{pick up the pen}
        \item[] 7. write names on the cards 
    \squishend
    &
    good plans shouldn't include redundant steps	
    &
    \edittype{remove node} \editloc{`pick up the pen'}	
    &
    \edittype{remove node} \editloc{`pick up the pen'}
    &
    1 & 1 & 1
    \\ \hline



    wrong order
    &
    \reallysquishlist 
        \item[] 1. \red{leave home and get in car}
        \item[] 2. remem. destination address 
        \item[] 3. \red{look around for the car}
        \item[] 4. walk towards the car
        \item[] 5. open the car door
        \item[] 6. sit down in the car
        \item[] 7. put on the seatbelt
    \squishend 
    &
    you wouldn't look for something you're already with	
    &
    \edittype{reorder edge} \editloc{between `$\langle$ leave home and get in car , look around for the car $\rangle$'}	
    &
    \edittype{remove node} \editloc{`look around for the car'}
    &
    0 & 0 & 0
    \\ \hline
    
    \end{tabular}
    \caption{Some examples of the data points and model predictions. \editm takes the form: \edittype{\texttt{<EDIT TYPE>}} over \editarg{\texttt{[<ARG>]}} at \editloc{\texttt{<LOCATION>}} The dataset contains partial order points as well, but they are omitted here for simplicity.
    }
    \label{tab:sample-predictions-interscript}
\end{table*}

\subsection{Task}

\begin{figure}[!ht]
\centering
{\includegraphics[width=0.8\columnwidth]{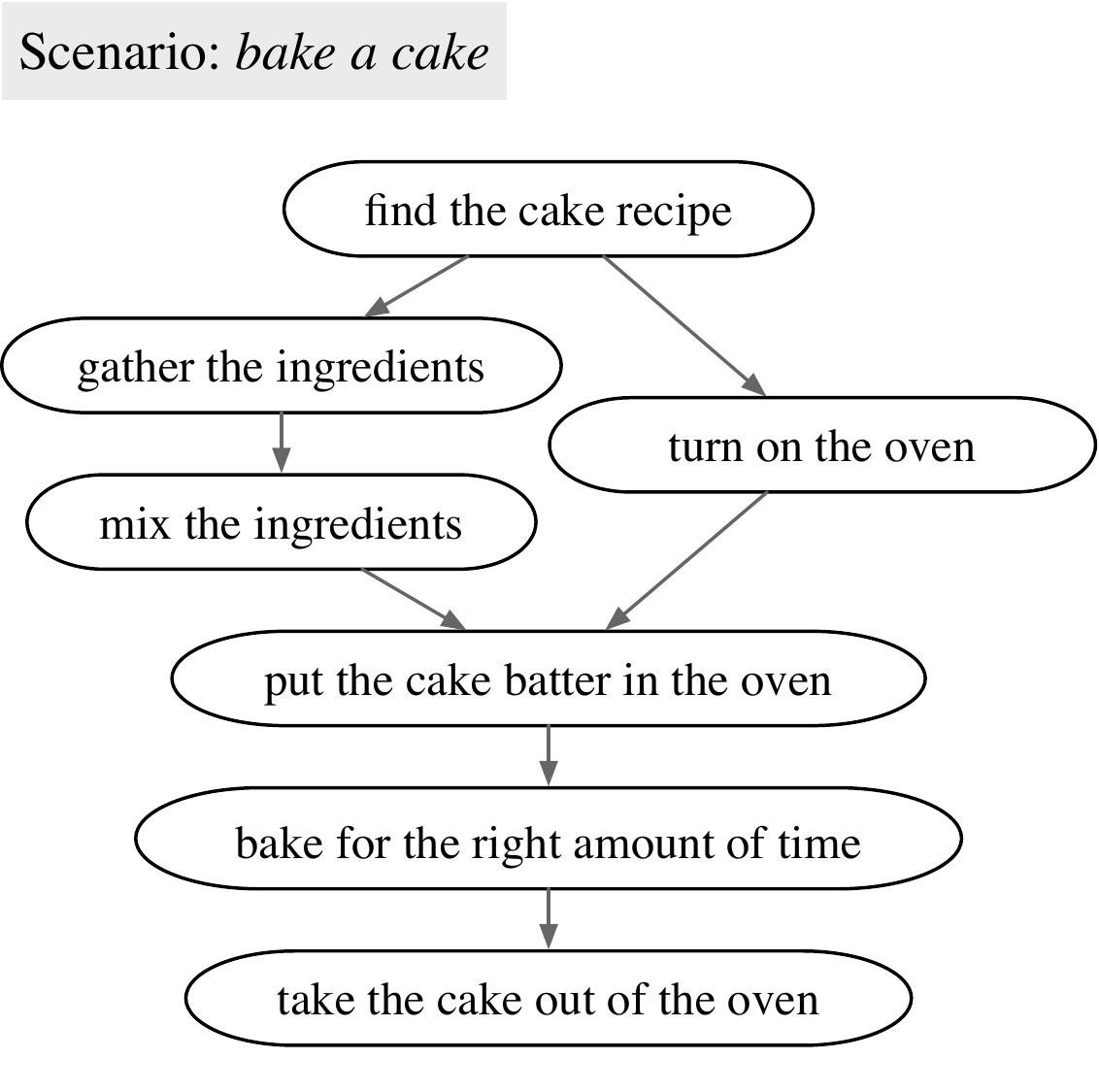}}
\caption{An example of a \textit{script} in \citet{proscript}. In a script generation task, models take the goal as the input and generate a (possibly) partial-order graph, which consists of essential steps and their ordering.}
\label{fig:proscript_orig_example}
\end{figure}

We instantiate our framework for the task of {\it script generation}.
Formally, the script generation task \cite{proscript} takes as input a scenario and generates a script $G(V, E)$, where $V$ is a set of essential events $\{v_1, ... v_i, ... v_{|V|}\}$ and $E$ is a set of temporal ordering constraints between events $\{e_{ij}\}$ which means that the events $v_i$ must precede the event $v_j$ ($v_i \prec v_j$). Partial ordering of events is possible, e.g., you can wear a left sock and a right sock in any temporal order. To solve this task, script generation models are required to \textit{generate} events ($V$) and predict the edges ($E$) jointly. See Figure \ref{fig:proscript_orig_example} for an example.

\proscriptgen  \cite{proscript} is a recently released model that, given a goal, generates $V$ and predicts the edge structure $E$ jointly. It is based on the \tfxxl model (11B parameters) and generates the script as a graph in DOT format. The authors report that the DOT format is always valid at inference time and that $V$ and the graph structure are generally of high quality. They characterize the graph edits required to correct a generated script (such as removing a node, adding a node, changing edge order, etc.). Mechanical Turk workers could repair most of the generated scripts within a few edits (typically an edit distance of 5) - we further validate this in Appendix \S\ref{sec:subsec:study}. This makes for an attractive use-case for interactive learning because the generated content from the model is repairable through user feedback. 

\subsection{Feedback Data Collection \label{dataset-collection}}

To train the corrector $G$, as well as evaluate our approach, we collected a set of $(x,fb,y)$ tuples using crowdworkers,
where $x$ is a possibly erroneous script generated by \proscriptgen, $fb$ is general feedback about the error (if any), and $y$ is the corrected script. In practice, crowdworkers specified the edits to $x$ to create $y$ (using simple graph operations we can generate \ansm from \editm -- see Table \ref{tab:sample-edit-to-output-graph} for an example). We collected 1542 tuples of data, randomly splitting it into 843 train, 154 validation, and 545 test points. Examples of the resulting dataset are shown in Table~\ref{tab:sample-predictions-interscript}.

\subsection{Training the Corrector Model}

We initialize $\theta$ with a checkpoint from the text-to-text pre-trained T5 transformer \cite{Raffel2020ExploringTL} and fine-tune on our dataset.  We use the default hyperparameters (including the Adafactor optimizer) in the T5 library.\footnote{https://github.com/google-research/text-to-text-transfer-transformer} We fine-tune a \tfxxl model for the main results, fine-tuned for 5,000 steps (batch size 8), selecting the checkpoint with the highest validation score (usually the final step).
To implement the memory \memorym, we use a BERT-based Sentence Transformer to encode $x$ ~\citep{reimers-2019-sentence-bert}, and use cosine distance with a threshold of 0.9 to find a matching key $\ques_m$. We leave the investigation of more complex retrieval functions (e.g., using attention mechanism to future work.)

\section{Experiments}

We empirically evaluate two questions:
\squishlist
\item[RQ1.] \textbf{How well does \ours interpret NL feedback?}  Specifically, we measure how well \ours can translate general feedback \fbm from a user into the correct repair edit on an imperfect script $x$. The main focus of RQ1 is to test the performance of \correctorm in the pipeline (Fig. \ref{fig:method})
\item[RQ2.] {\bf How well can \ours learn from prior mistakes?}  We make the same measurement, but using feedback \fbm recalled from similar, previous examples. The main focus of RQ2 is to test the performance of \memorym and $\Omega$.
\squishend

\paragraph{Metrics} To compare the gold edit \editgoldm and the generated edit \editgeneratedm, we use standard metrics used to evaluate generated text. 
We report the following metrics:
\squishlist
\item \textbf{Exact match}: \exact gives a score of 1 if \editgoldm is equal to  \editgeneratedm and 0 otherwise. 
\item \textbf{Generation metrics}: We report standard generation metrics \bleu~\citep{papineni2002bleu} and \rouge~\citep{lin2004rouge} to account for similar but not exact matches.
We use the implementation released in the metrics package of the GEM-benchmark~\citep{gehrmann2021gem}.\footnote{\url{https://github.com/GEM-benchmark/GEM-metrics/}}
\squishend
We report these metrics over the entire edit: \exact, \bleu, \rouge. The components of \editm broadly follow a template: \texttt{<EDIT TYPE> over [ARG] at <LOCATION>} (see Table \ref{tab:sample-predictions-interscript}). This allows comparison of the location or edit type in \editgoldm and \editgeneratedm: \exact$_{loc}$, \bleu$_{loc}$, \rouge$_{loc}$ and  \exact$_{type}$, \bleu$_{type}$, \rouge$_{type}$

\paragraph{Baseline} As baseline, we train a model that does not use any feedback~(we call this, \nfb) and is trained only with \textit{input = erroneous script} and \textit{output = edit}. 
The language model used in this baseline and \ours is the same~(\tfxxl), allowing a meaningful comparison.


\subsection{RQ1: How well does \ours interpret NL feedback?}
\label{sec:mainresult1}
To measure how well the graph corrector \correctorm learns to interpret NL feedback, we provide oracle feedback to \ours, and we call this $\oursoraclefb$.  
Table \ref{tab:main_results} shows that $\oursoraclefb$ learns to react to the feedback, \ncr{as indicated by a sharp increase in both the exact match scores and automated metrics}. 
Further, we note that the model is good at identifying the error type that the feedback indicates. Still, it is difficult for the model to localize the error in the graph, probably because the location is not explicitly mentioned in the feedback, and the model must infer it.


\begin{table}[]
\centering

\addtolength{\tabcolsep}{-2.5pt}  
\begin{tabular}{@{}lrrrrr@{}}
\toprule
  & $EM$ & $EM_{loc}$ & $EM_{type}$ & \bleu & \rouge \\ \midrule
\nfb   & 3.5  & 9.7 & 30.4 & 21.7 & 39.0       \\
$\oursoraclefb$ &  \textbf{38.6}  &  \textbf{45.8} & \textbf{69.3} & \textbf{54.2} & \textbf{70.6}         \\ \bottomrule
\end{tabular}
\caption{{\bf Interpreting NL Feedback:} Correctness of Predicted Edits .... Given an erroneous script $x$, and general feedback $fb$ 
\underline{from the user}, FBNet perfectly predicts the specific repair edits 38\% of the time ($EM$, exact match) -
an order of magnitude better than a baseline \nfb{} predicting the repair
from $x$ alone. $EM_{loc}$ and $EM_{type}$ compare just parts of the edit
sequences (locations/types of the required edits, respectively), while
\bleu and \rouge are softer matching metrics. 
}
\label{tab:main_results}
\addtolength{\tabcolsep}{2pt}  
\end{table}

\paragraph{How consistently does \ours interpret similar feedback?}
In $\sim$15\% of the data points, multiple \fbm can lead to the same $(\ques, \ans)$ pair. 
\ours is expected to behave consistently for such re-phrasings of \fbm. The model consistently produces exactly the same \ansm for \fbm re-phrasings $\sim$ 60\% of the time. Furthermore, we observe majority agreement as the number of \fbm re-phrasings for a $(\ques, \ans)$ pair increases. In our analysis, a large proportion of the inconsistent edits occur because different \fbm phrasings prompt the model to generate slightly different, but semantically similar edits: see Table \ref{tab:one_error_multiple_feedbacks_2_fb_example} for an example.

\begin{table}[!h]
    \centering
    \begin{tabular}{|p{0.45\columnwidth}|p{0.45\columnwidth}|}
    \hline
    feedback & predicted edit \\
    \hline
    The feedback is if a person is going to open a book, they need to choose one first & insert node `choose a book to read' before `open the book' \\ 
    \hline
    The feedback is you can't open something you're not holding & insert node `get the book out of the bag' before `open the book' \\
    \hline
    \end{tabular}
    \caption{Multiple feedbacks for the same (\quesm, \ansm). Here, \quesm is: You are given a plan to read to child. decide which books to read, open the book, read the book to the child, turn the pages ... . \editm is \textit{insert node `pick a book off the shelf' before `open the book'}}
    \label{tab:one_error_multiple_feedbacks_2_fb_example}
\end{table}


\paragraph{How well can \ours handle wrong feedback?} 
While the ability to react to feedback is a desired trait for \ours, we also want to ensure that the performance of \ours is proportional to the quality of feedback.
This will ensure that \ours can act faithfully in settings where the feedback might be potentially misleading.
We investigate this question by identifying lexically similar scripts but irrelevant feedback from the training set for each test example. Note that our setup easily allows us to test this hypothesis since the train/test/val splits were carefully designed to ensure no overlap between the examples. Thus the feedback from one example will typically not apply to another example. 
We find that with irrelevant feedback, the performance of \ours drops to 3\%.
This shows that \ours is sensitive to the quality of feedback, and no feedback is better than misleading and irrelevant feedback.

\paragraph{How well does \ours perform across error types?} $\oursoraclefb$ gets the highest performance ($EM$ 63.0\%) on wrong-step error type where \fbm typically contains negative words that signal the error type, and the model learns to localize the error node. One of the most challenging error types is partial order removal or addition ($EM$ 10.5\%). This can be attributed to the challenging localization involving multiple nodes that participate in a partial order. The lowest-performing is the missing step error type ($EM$ 2.73). 
The reason for this low $EM$ score is that the edit \ncr{must \textit{generate} the missing node, and $EM$ undercounts the correctness of the generated text.} Other metrics such as \rouge are much higher validating that the model performs well on this error type. Section \secref{sec:perfbyerrortype}  Table~\ref{tab:embytype} breaks down the performance of \ours by error type.


    
\subsubsection{Error analysis}

We randomly sampled 50 instances from the test set where the model generates an incorrect edit~(i.e., $EM=0$).  
Our goal is to understand the typical errors made by the model and use the analysis to calibrate the findings in Table~\ref{tab:main_results}.

\squishlist    
\item \textbf{Lexical variation (36\%)} Exact match underestimates the performance of our model (as the task involves generation). We find that more than 35\% of the predicted edits are semantically similar (typically lexical variation) to the reference gold edit. Some examples include: insert node \textit{picking a book...} vs, insert node \textit{choosing a book to read}. Another kind of example is the model suggesting swapping the order of \textit{edges A and B} while the reference edit swaps \textit{edges B and A} - but both of these are equivalent.
    
    

\item \textbf{Challenging feedback (24\%)} 
This type of error occurs when the model fails to interpret a feedback because it is difficult to interpret e.g., the feedback is expressed abstractly.
For example, for the goal ``go to locker room,'' the generated script repeats the step ``walk to the locker room.''. However, the feedback is `you can't go where you already are', and \ours generates the edit ``reorder edge between `$\langle$ walk towards the locker room , walk to the locker room $\rangle$' '' , failing to interpret the feedback.


\item \textbf{Error not localized (20\%)}
In about 20\% of the failures, \ours fails to localize the error given the feedback.
For example, consider the erroneous input script about the goal \textit{buy an xbox}:
    \textit{1. go to the store 2. talk to the cashier 3. make the transaction  4. get the receipt 5. load the video game into the car 6. get into the car 7. take xbox home}   
The feedback is \textit{after a person makes a transaction, they then head to their car}. 
The expected edit is: \textit{insert node `walk to the car' after `get the receipt'}, but the predicted edit \textit{insert node `get into the car' after `make the transaction'} does not correctly identify the erroneous node. The feedback points to making a transaction, but it also involves getting the receipt. 

\item \textbf{Alternative answers (16\%)} 
We also encounter cases where there are multiple ways to correct a script. For example, an edit can be expressed as \textit{insert node `X' before `step 4'} or \textit{insert node `X' after `step 3'}. This comprises $\sim$ 16\% of the errors.
\squishend

In $\sim$32\% cases, the model generates a correct edit that differs from the gold. 
Extrapolating this performance under-counting to the entire test set, the accuracy of \ours in Table \ref{tab:main_results} would increase to $\sim$70\% (+32\%). 

\subsection{RQ2: How well can \ours learn from prior mistakes?}
\label{sec:mainresult2}



\begin{table}[!h]
\centering
\addtolength{\tabcolsep}{-4pt}
\begin{tabular}{@{}lrrrrr@{}}
\toprule
               & $EM$    & $EM_{loc}$    & $EM_{type}$  & \bleu & \rouge  \\ \midrule
$\;$ \nfb           & 6.94    &  15.3         &  34.7        & 24.1     & 44.2       \\
$\;$ \ours          & 16.72   &  20.9         &  56.9        & 32.5     & 48.5         \\ \
$\;\oursoraclefb$    & 22.2    &  27.8         &  72.2        & 44.6     & 65.8          \\ \bottomrule
\end{tabular}
\caption{{\bf Learning from prior mistakes:} On the reuse dataset, given an erroneous script $x$, and feedback $fb$ 
\underline{recalled from similar, prior examples}, FBNet perfectly predicts the specific repair edits 16.7\% of the time
(or 20.9\% the edit location and 56.9\% the edit type), a promising start to
learning from prior mistakes.}
\label{tab:rq2results}
\end{table}




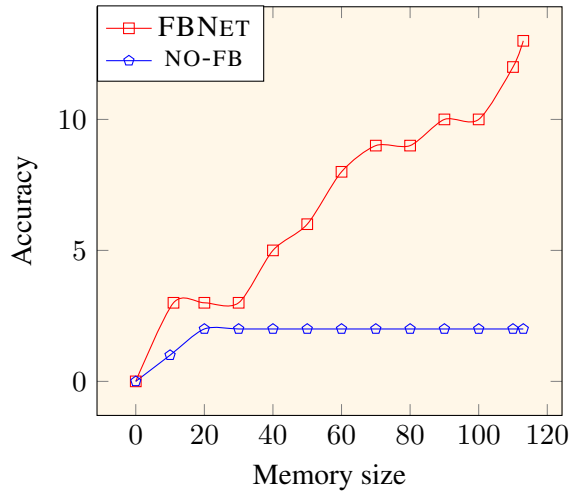
\begin{figure}
\centering
    \begin{tikzpicture}
        \begin{axis}[
        	xlabel=Memory size,
        	ylabel=Accuracy, 
        	width=1\columnwidth,height=7cm,
        	axis background/.style={fill=cosmiclatte},
            legend style={at={(0.0,.91)},anchor=west}]
        \addplot[color=red,mark=square, smooth] coordinates {
        	(0,0)
            (11,3)
            (20,3)
            (30,3)
            (40,5)
            (50,6)
            (60,8)
            (70,9)
            (80,9)
            (90,10)
            (100,10)
            (110,12)
            (113,13)
};
        \addplot[color=blue,mark=pentagon, smooth] coordinates {
        	(0,0)
            (10,1)
            (20,2)
            (30,2)
            (40,2)
            (50,2)
            (60,2)
            (70,2)
            (80,2)
            (90,2)
            (100,2)
            (110,2)
            (113,2)
};
        \legend{\ours, \nfb}
        \end{axis}
        \end{tikzpicture}
\caption{Performance on unseen examples (number of correct data points) improves as memory size grows. \nfb~ baseline performance remains static. \ncr{Note that accuracy is evaluated using exact match, and thus is a lower bound on the actual equivalence as exact match might miss rephrasings.}}
\label{fig:continuous-learning}
\end{figure}

Section~\secref{sec:mainresult1} shows that the corrector \correctorm can utilize user-supplied feedback to fix an incorrect structure.
\ours combines \correctorm with a memory \memorym of feedback, allowing us to leverage past feedback on new examples.
This section presents a setup where feedback on previously seen inputs is used to fix new, unseen examples.


To investigate this setting, we create a new test set, called the \textit{\iset} or \textsc{iset}. To create it, we randomly sample 72 test points (referred as \isetMother or \textsc{iset-source}) and perturb them linguistically to generate \iset (also referred as \textsc{iset}).  
The perturbations are performed on the salient entities in the script, including (i) linguistic perturbation on $\sim$20\% samples (e.g., box $\rightarrow$ carton, package) and (ii) the relatively harder analogical perturbation on the remaining $\sim$80\% samples (e.g., bus $\rightarrow$ train, and how to lift blinds $\rightarrow$ how to open oven door because the event structure is analogical). The \editm to the original script also applies to the substituted script. We ensured that the perturbations did not introduce additional errors in the substituted script. 
This ensures that the \iset now contains similar examples to the original test set, a condition that our original splits do not satisfy. There are a total of 72 data points in \iset.

\SetKwInput{KwGiven}{Given}
\SetKwInput{KwInit}{Init}
\begin{algorithm}[tb]
\SetAlgoLined
\KwGiven{\ours, \memorym, \retm}
\KwGiven{Set $\{$\textsc{iset} $\cup$ \textsc{iset-source}$\}$ of $N$ queries.}

\For{$i \gets 1, 2, \ldots, N$}{
\tcc{\small{Check memory for feedback}}
$\bar{\fb}_i =\ \ret(\ques_i, \memory) $\;
\tcc{\small{Get corrected structure from \ours. $\bar{\fb}_i$ can be empty.}}
$\ans_i =\ \ours(\ques_i, \bar{\fb}_i) $\;
\tcc{\small{Get user feedback}}
$\fb_i = $ User feedback on $\ans_i$\;
\tcc{\small{Grow memory with new feedback}}
Write $\fb_i$ to \memorym
}
\caption{\ours inference on a stream of inputs with growing memory}
\label{alg:alg-iset}
\end{algorithm}

\paragraph{Continually learning using a memory of errors}
Examples in \iset~ are randomly mixed with the original test set.
This combined test set of queries $Q$ is then evaluated using our setup as shown in Algorithm~\ref{alg:alg-iset}.
Intuitively, \iset~ allows us to simulate a setting where the system has been deployed in the wild, and end-users can query.
Algorithm~\ref{alg:alg-iset} runs the memory-based inference described in Section~\secref{sec:method} (Figure~\ref{fig:method}).
As the system is run through the stream of queries, we expect that i) the overall performance of the system will be better than no feedback, as some of the examples in the interaction set will provide meaningful feedback, and ii) the \textit{running} performance of the system will improve with growing memory: the probability of relevant feedback being present for an unseen example increases with time, thus boosting the performance.

Our experiments show that \ours meets both these expectations. First, Table~\ref{tab:rq2results} shows that retrieved feedback improves over no feedback by 10 points (exact match) and similarly in terms of \bleu and \rouge scores, respectively. Further, Figure~\ref{fig:continuous-learning} shows a graph confirming that \ours can improve continuously as memory grows.

%




\section{Scope}

In principle, we could apply \ours to any task that satisfies the assumptions (\S \ref{sec:assumptions}). However, our approach has some limitations in practice, several of which merit further detailed follow-up work. 

\squishlist
\item \textbf{On Assumption A1:} We assume that the output of \basemodelm is  repairable. Such an assumption is only possible for models that generate mostly correct outputs and have errors that are easy to highlight for humans. In practice, this implies that our approach will most efficiently work in conjunction with modern language models \citep{foundational-models-stanford} that are shown to be syntactically correct in form, but can produce output that lacks commonsense \citep{BenderKoller2020FormAndMeaning}, making their output repairable.

\item \textbf{On Assumption A2:} Having reusable, general feedback is costly and requires careful instructions to collect from general users and crowdworkers (e.g., we asked the crowdworkers how they would explain the model error to a five-year-old). As the domain of the task becomes more specialized, such as database query generation \citep{elgohary2021nl} or code correction \citep{Yasunaga2020Bifi-v0-PercyLiang},  collecting data to train \correctorm becomes difficult. Systems that produce structured explanations are better suited to our model (see \citet{Wiegreffe2021TeachMT} for an overview), rather than specialized domains that require expert users to provide feedback (e.g., in database query generation).
\item \textbf{Consistent memory:} We show in Section~\secref{sec:mainresult1} that \ours is sensitive to the appropriateness of the feedback. However, adversarial or incorrect feedback could pollute the memory and possibly make it inconsistent. There has been some recent work to ensure consistency of beliefs of a model \citep{Kassner2021BeliefBankAM}, and more effort is required in this direction to apply to more complex settings like ours.

\item \textbf{Using multiple feedbacks:} \retm can be enhanced with more complex attention mechanisms that aggregate from multiple relevant memory entries and possibly generalize them. We conducted an initial experiment using attention and found that we would need a larger dataset to train \retm effectively.
\squishend

Advancements in these directions would further increase the applicability of \ours. Still, there are several applications \citep{Wiegreffe2021TeachMT} where our approach would currently apply in principle, or is easy to set up.

\section{Summary}

Our goal is to create a system that can continuously improve the structured output of a model.
Our approach is to train an error correction model that uses natural language (NL) feedback to correct errors in that output.
We have presented the first step towards this goal, showing that an error correction module can learn to interpret NL feedback successfully, resulting in 40\% fewer errors in script generation. 
We have also described ongoing work on the next step, namely adding a memory layer where human feedback is stored and later retrieved efficiently.
Together, these offer a possible path to systems that can continuously improve their output over time, with progressively less feedback and without retraining.

\section*{Acknowledgments}
This material is partly based on research sponsored in part by the Air Force Research Laboratory under agreement number FA8750-19-2-0200. 
The U.S. Government is authorized to reproduce and distribute reprints for Governmental purposes notwithstanding any copyright notation thereon. 
The views and conclusions contained herein are those of the authors and should not be interpreted as necessarily representing the official policies or endorsements, either expressed or implied, of the Air Force Research Laboratory or the U.S. Government.
We would like to thank Google for providing the TPU machines for conducting experiments.

\bibliographystyle{acl_natbib}
\bibliography{acl2021fixed}

\newpage
\clearpage

\section{Appendix}

\subsection{Initial study on the errors of \basemodelm (\proscript)}
\label{sec:subsec:study}
On \proscript's test set, we performed inference using the released checkpoint (both GPT-2 and T5-XXL based model). We then randomly sampled 30 generated graphs and manually wrote feedback for them. Similar to \citet{proscript}, we found that the model makes repairable mistakes (leading to assumption A1 being satisfied). Further, we found there instances where a general principle feedback applies across more than one instances (e.g., you have to be near something to use it). (see Table \ref{tab:study-sample-feedback}).

\begin{table}[!h]
\small
    \centering
    \begin{tabular}{|p{0.45\columnwidth}|p{0.45\columnwidth}|}
    \hline \textbf{What was the error} &  \textbf{General principle feedback} \\ \hline
    Script was missing the step of not turning off the alarm after waking up & People don't leave their alarms ringing all day.\\ \hline
    Script mentioned coming to the doorway and passing through it & One cannot walk through the doorway without opening the door first. \\ \hline
    Script tells that getting in car and drive in zoo can be done in any order & People must get into a vehicle, before driving to any place. \\ \hline
    Script is looking for a butterfly after placing it & You don't need to look for a butterfly if it's already in a container.\\ \hline
    \end{tabular}
    \caption{Sample error and the corresponding general principle feedback that could, in principle, repair the model output.}
    \label{tab:study-sample-feedback}
\end{table}

On an average, there were about two mistakes present in the graphs. Often, the error was that the script was using an entity before having it (e.g., write on the paper comes before the node find the paper or reach for the paper). 
Thus, there seems to be a possibility of applying similar feedback to more than one example.
We also found some cases where the script might have to be changed to adapt to special cases. 
For example, for a script \textit{visit Disneyland}, an event \textit{obtain a visa} might be required for some users.
We believe the original ProScript dataset aims to generate widely applicable scripts and grounded in commonsense; rather than cover all possible outcomes. 


On the surface, the generated scripts were of good quality. However, a closer look at the mistakes revealed that most of them could be attributed to the model lacking basic commonsense. 
For example, Figure \ref{fig:intro-example} shows a typical mistake the model makes. 
This underscores the gap between the syntax and semantic correctness of machine-generated output in the context of automatic script generation. This observation is in-line with other NLP tasks \citep{BenderKoller2020FormAndMeaning} that distinguish the success of recent models on the correctness of form rather than the far-from-over goal of understanding of meaning.


\subsection{Data collection}

\begin{figure}[!ht]
\centering
{\includegraphics[width=0.9\columnwidth]{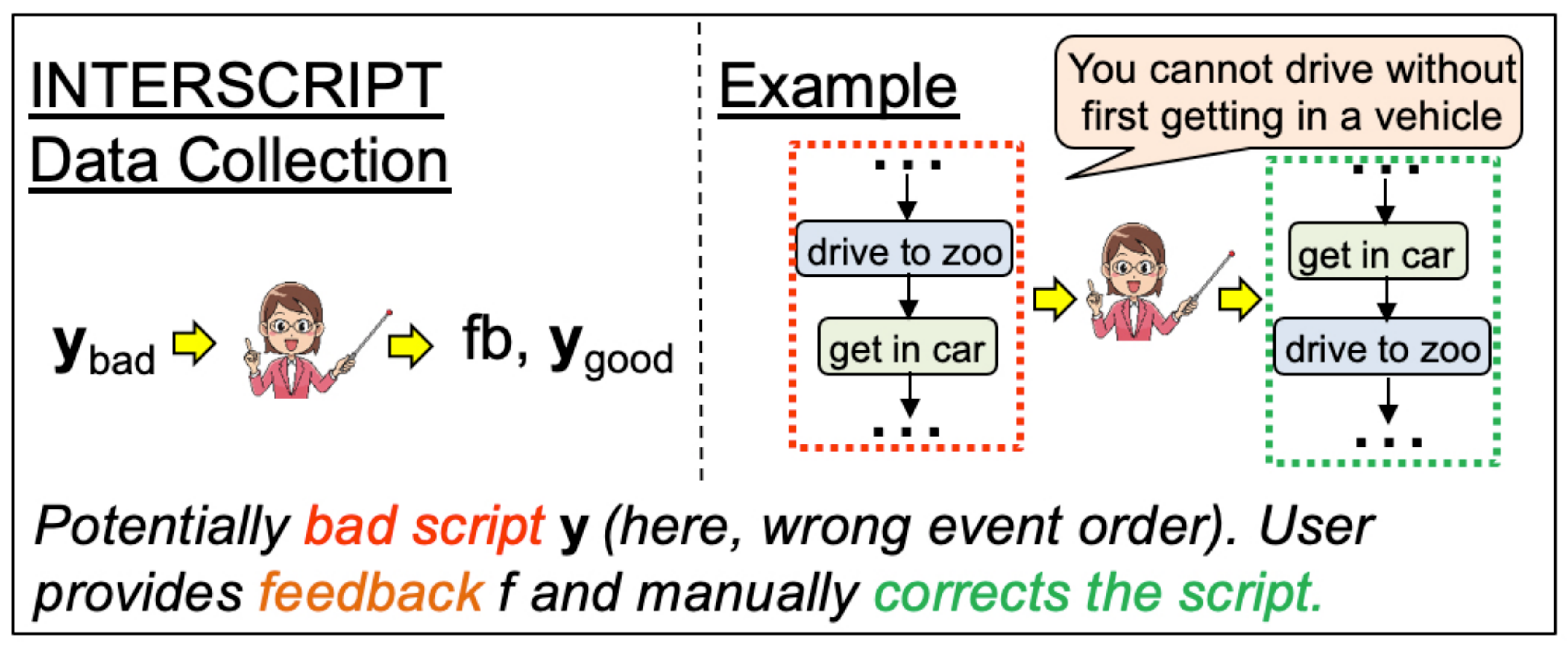}}
\caption{A broad overview of the annotation process. For actual annotation task (including the M-turk task template), see our code repository.
}
\label{fig:mturk-idea}
\end{figure}

An average user could point out mistakes in the generated scripts, as a majority of the errors in generated scripts are caused by a lack of basic commonsense~(\S{\ref{sec:subsec:study}}).
Consequently, we designed a Mechanical Turk task to provide feedback on mistakes.  A broad overview of the annotation process is shown in Figure \ref{fig:mturk-idea}. 

\paragraph{Annotation}
Now we discuss our crowdsourcing setup to collect the data. To maximize the opportunity to get more feedbacks for a predicted script, we filtered a subset of the test set in ProScript where the human evaluated graph edit distance was likely to be high (i.e., there were likely to be more errors). The ProScript authors released the graph edit value for the set of test set samples they evaluated. We performed inference using their released \proscriptgen model on those data points with high graph edit distance value ($\ge 8$). With this we collected about 400 (predicted graph, expected gold graph) tuples. The ProScript paper describes that their expected gold graph is also imperfect and might contain about 20\% noise. 
Nevertheless, having the gold reference graph guides and constrains an annotator about the common script for a scenario rather than the wide open space of solving the task using multiple potentially correct scripts. (e.g., one could go to a zoo without driving the car by hiring a taxi and then they won't need to drive or park the car). 
As mentioned in \S{\ref{sec:subsec:study}} our annotation process must focus on scripts that are widely applicable and grounded in commonsense.

\begin{figure*}[]
\centering
{\includegraphics[width=1.0\linewidth]{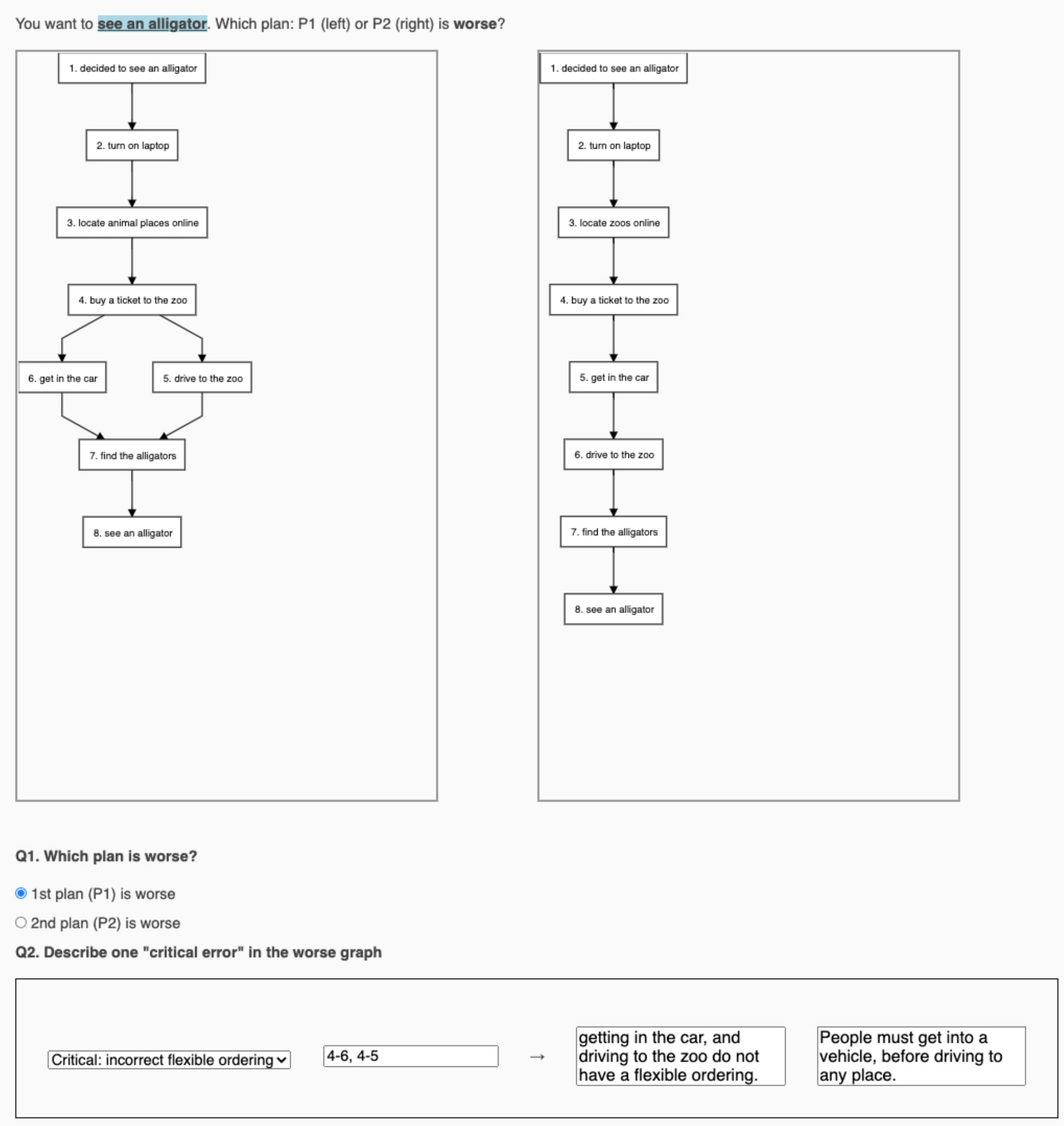}}
\caption{The mechanical turk page for annotation. We show the generated and the expected ProScript gold reference. The annotator must answer which script is worse and why. They must point out an egregious mistake (and not any trivial errors that have minor grammatical errors), and annotate: the error type (missing step, wrong step, wrong order, wrong partial order), localize the error by providing the node or edge id, and give feedback why it is wrong, and finally to gather the general principle behind the feedback they are asked to explain the feedback to a five-year-old.}
\label{fig:mturk-page}
\end{figure*}

The annotators are shown the model-generated and expected gold (reference) scripts, and are required to answer which script is worse and why.
It is possible that the gold script is marked as worse.
However, we later post-process and remove such cases, as our focus is to get errors on the generated scripts and not the manually created scripts.
The annotators must point out an obvious mistake (\eg an event or an edge that does not follow commonsense).
They were asked to ignore grammatical and fluency errors, and focus on critical errors of four types:
\squishlist
\item \textbf{Wrong ordering:} the order in the sequence of steps is not correct (\eg wearing shoes is described before wearing socks).
\item \textbf{Flexible ordering:} some steps can be done in a flexible order (\eg you can wear left sock or right sock first). A good script captures such flexibility.
\item \textbf{Missing critical steps:} a bad script might have missed critical steps (\eg the script can say: ``wait for a plane'' followed by ``get off the plane'' -- here an obvious step ``get on the plane" is missing) . There is no strict definition for a critical step, so the annotators were instructed to use their commonsense judgment.
\item \textbf{Wrong step:} a bad script might have irrelevant and wrong steps (\eg the script describing ``go to a party" might describe an irrelevant step such as read a book, open a book, etc.).
\squishend

For every data point, the annotators were asked to answer the following: 
\squishlist
    \item Explicit feedback type-1: the error type (missing step, wrong step, wrong order, wrong partial order)
    \item Explicit feedback type-2: localize the error by providing the erroneous node or edge id
    \item Implicit feedback type-1: give feedback in a few words, explaining the error
    \item Implicit feedback type-2: An explanation of the error that would potentially make sense to a five-year-old. Such an explanation of the feedback helped gather the general principle that is violated, and is targeted in the feedback.
\squishend
Fig. \ref{fig:mturk-page} shows a sample of our Mechanical Turk task. Annotators were required to list only one critical error that they believe was most important.
Each data point is annotated by three annotators, adding some diversity in the errors. 
The annotators were paid \$15 an hour. The annotators were English speaking crowdworkers on Mechanical Turk from USA. 
The average time for completion of one script was 2 minutes. 

\begin{table}[!h]
  \small
    \centering
    \begin{tabular}{|p{0.2\columnwidth}|p{0.1\columnwidth}||p{0.58\columnwidth}|}
    \hline \fbm type &  count & example \\ \hline
explicit \fbm type-1 & 1,553 & Remove node `put the shirt on' \\ \hline
explicit \fbm type-2 & 1,553 & The following step is not right: put the shirt on \\ \hline
implicit \fbm type-1 & 1,553 & It tells you to iron your shirt while it's still on your body. \\ \hline
implicit \fbm type-2 & 1,553 & If you hold a hot iron against the clothes you're currently wearing, you'll get terrible burns. \\ \hline
total & 6,212 & \url{https://anonymous.4open.science/r/interscript/data.json} \\ \hline
    \end{tabular}
    \caption{Dataset statistics. In this paper, we use the hardest feedback (implicit-feedback-type-2). This example is from the input script: input script for the following table (goal: press the wrinkles out) = 
                1. put the shirt on,
                 2. find place to press,
                 3. grab iron from drawer,
                 4. place iron on shirt,
                 5. wait for iron to heat up,
                 6. use iron to smooth out wrinkles,
                 7. press the wrinkles out}
    \label{tab:dataset_statistics_by_feedback_type}
\end{table}

We measured the agreement on labels (which graph is worse), and on explicit feedback type-1 and type-2. It was difficult to measure agreement on implicit feedback because it is not easy to perform binary comparison on the generated text without accounting for linguistic variations. On the labels, the Fleiss Kappa agreement was 0.90 (almost perfect agreement) and on explicit feedback the agreement was 0.75 Fleiss Kappa (substantial agreement). This also shows that there is some diversity in what the users perceive as a serious mistake in \badgm. 

Eventually, we compiled these annotations into a dataset of 1,553  4-tuples where each entry consists of (explicit feedback type-1, explicit feedback type-2, implicit feedback type-1, implicit feedback type-2). 

\begin{table*}[!h]
    \centering
    \small
    \begin{tabular}{|p{0.25\textwidth}|p{0.15\textwidth}|p{0.15\textwidth}||p{0.25\textwidth}|}
    \hline 
    \quesm &  \fbm & \editgoldm & \ansm  \\ \hline
    
    \reallysquishlist 
        \item[] 1. ... 
        \item[] 2. \red{set alarm for early morning}
        \item[] 3. \red{get out of bed}
        \item[] 4. prepare for yoga 
        \item[] 5. go to the bathroom 
        \item[] 6. do yoga 
        \item[] 7. do yoga in the morning 
    \squishend
    & 
        \reallysquishlist 
            \item[] People don't leave their alarms ringing all day. 
        \squishend
    &
    insert node `wake up and turn off alarm' before `get out of bed'
    &
    \reallysquishlist 
        \item[] 1. ... 
        \item[] 2. set alarm for early morning
        \item[] 3. \green{wake up and turn off alarm}
        \item[] 4. get out of bed
        \item[] 5. ... 
        \item[] 6. ... 
        \item[] 7. ... 
    \squishend
    
    \\ \hline

    \reallysquishlist 
        \item[] 1. ... 
        \item[] 3. put on shoes ... 
        \item[] 5. open the door
        \item[] \red{6. drive to the train station} 
        \item[] \red{7. get into the car} 
        \item[] 8. reach the train station 
    \squishend
    & \reallysquishlist 
            \item[] You must get into a vehicle, before driving to any place. 
      \squishend
    &
    reorder edge between `$\langle$ drive to the train station, get into the car $\rangle$'
    & 
    \reallysquishlist 
        \item[] 1. ... 
        \item[] 3. ... 
        \item[] 5. open the door 
        \item[] \green{6. get into the car.} 
        \item[] \green{7. drive to the train station}
        \item[] 8. ... 
    \squishend
    \\ \hline
    
    \reallysquishlist
        \item[] 1. ... 
        \item[] 3. pick up the butterfly
        \item[] \red{4. put the butterfly in container} 
        \item[] \red{5. look for a butterfly ...}
        \item[] 6. Take the butterfly home ... 
    \squishend
    & 
    \reallysquishlist 
        \item[] You don't need to look for a butterfly if it's already in a container. 
    \squishend
    &
    remove node `look for a butterfly'
    & 
    \reallysquishlist 
        \item[] 1. ... 
        \item[] 3. pick up the butterfly 
        \item[] \green{4. put the butterfly in container} 
        \item[] \green{5. Take the butterfly home}
        \item[] 6. ... 
    \squishend 
    \\ \hline
    
    \end{tabular}
    \caption{Task: Applying the graph edit to the bad script.
    }
    \label{tab:sample-edit-to-output-graph}
\end{table*}

\subsection{Model Output Examples}
Table \ref{tab:sample-edit-to-output-graph} provides some sample model outputs.

\section{Performance of \ours by error type}
\label{sec:perfbyerrortype}
\begin{table}[]
\centering
\begin{tabular}{@{}lr@{}}
\toprule
Edit type                       & EM\% \\ \midrule
Overall & 38.6 \\ \midrule
Add partial order exactmatch    & 10.5 \\
Add partial order type          & 44.7 \\
Missing step exactmatch         & 2.8 \\
Missing step type               & 65.5 \\
Remove partial order exactmatch & 0.0    \\
Remove partial order type       & 0.0    \\
Wrong ordering exactmatch       & 45.1 \\
Wrong ordering type             & 72.8 \\
Wrong step exactmatch           & 63.0   \\
Wrong step type                 & 78.6 \\ \bottomrule
\end{tabular}%
\caption{Performance of \ours by error type}
\label{tab:embytype}
\end{table}

Table~\ref{tab:embytype} breaks down the performance of \ours by error type.

\end{document}